\documentclass[11pt,a4paper]{article}
\usepackage[hyperref]{emnlp2020}

\usepackage{times}
\usepackage{latexsym}

\usepackage{microtype}

\usepackage[utf8]{inputenc}
\usepackage{tikz}
\usepackage{float}
\usepackage{booktabs}
\usepackage{graphicx}
\usepackage{multirow}
\usepackage{xspace}
\usepackage{textcomp}

\aclfinalcopy 
\def\aclpaperid{1428} 

\newcommand{\frozen}{\includegraphics[scale=0.067]{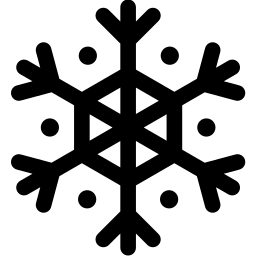}}
\newcommand{\unfrozen}{\includegraphics[scale=0.5]{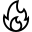}}
\newcommand{\untied}{\includegraphics[scale=0.5]{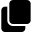}}
\newcommand{\tied}{\includegraphics[scale=0.4]{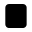}}
\newcommand{\dice}{\includegraphics[scale=0.02]{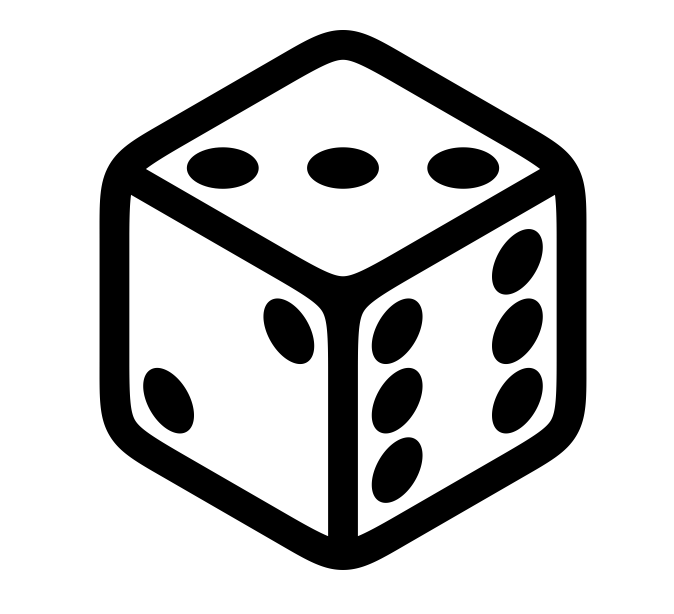}}
\newcommand{\train}{\includegraphics[scale=0.02]{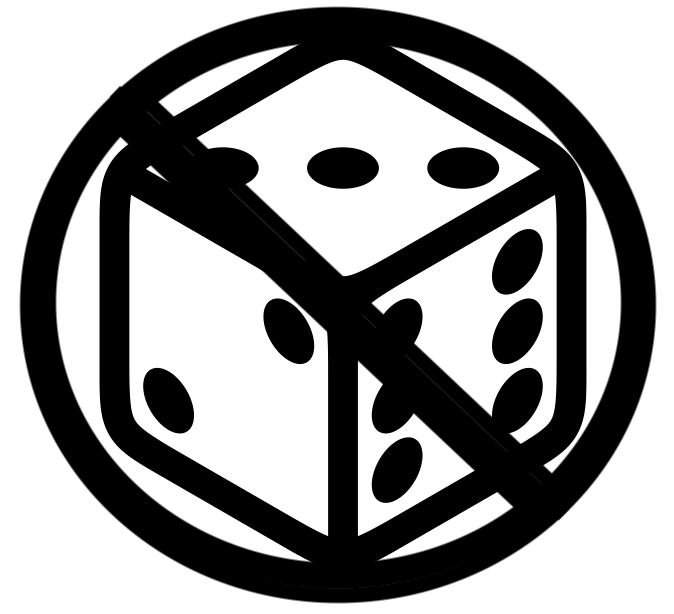}}
\newcommand{\halftrain}{\includegraphics[scale=0.02]{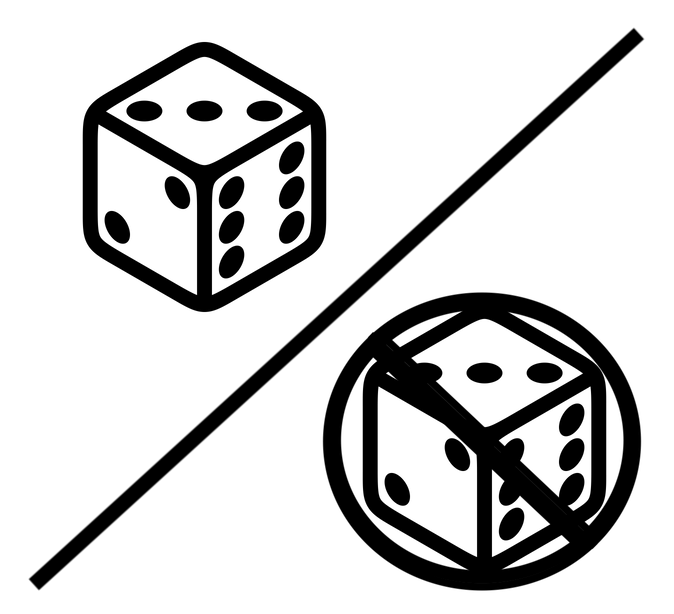}}

\newcommand{\tightparagraph}[1]{\textbf{#1}}
\newcommand{\tablescale}{0.9}

\title{Improving Low Compute Language Modeling with\\ In-Domain Embedding Initialisation}

\author{Charles Welch, Rada Mihalcea \and Jonathan K. Kummerfeld \\
Computer Science \& Engineering \\
University of Michigan \\
\texttt{\{cfwelch,mihalcea,jkummerf\}@umich.edu}
}

\date{}

\begin{document}
\maketitle
\begin{abstract}
    Many NLP applications, such as biomedical data and technical support, have 10-100 million tokens of in-domain data and limited computational resources for learning from it.
    How should we train a language model in this scenario?
    Most language modeling research considers either a small dataset with a closed vocabulary (like the standard 1 million token Penn Treebank), or the whole web with byte-pair encoding.
    We show that for our target setting in English, initialising and freezing input embeddings using in-domain data can improve language model performance by providing a useful representation of rare words, and this pattern holds across several different domains.
    In the process, we show that the standard convention of tying input and output embeddings does not improve perplexity when initializing with embeddings trained on in-domain data.
\end{abstract}

\section{Introduction}

Language modeling is an essential part of many NLP applications, including predictive keyboards, speech recognition, and translation.
Recent work has focused on
(1) small constrained datasets, such as the Penn Treebank \cite{ptb} and WikiText-103 \cite{wikitext103}, and
(2) vast resources with billions of words used to train enormous models with significant computational requirements \cite{gpt2}.
This leaves a gap: when a substantial amount of in-domain data is available, but computational power is limited.

We explore how initialising word embeddings using in-domain data can improve language modeling in English. 
Testing all valid configurations of weight tying, embedding freezing, and initialisation, we find that the standard configuration is not optimal when rare words are present.
Instead, the best approach is to initialise with in-domain data, untie the input and output, and freeze the input.

To understand this difference, we run a series of experiments to measure the impact of changing (a) the threshold for replacing rare words with a special symbol; (b) the source of data for initialisation; (c) the amount of training data for the language model; and (d) the hyperparameters for both the baseline and our proposed approach.
We find that the improvement comes from improved representation of rare words.
These findings are confirmed through experiments on four additional domains, with similar trends.

We also compare our approach to an n-gram language model and a large-scale transformer model. 
We find that if a large-scale transformer is inappropriate either for computational or modeling reasons, it is best to train an LSTM-based language model with as much data as possible and initialise the embeddings on all available in-domain data.

\section{Proposed Approach}\label{sec:ptb}

We propose initialising the language model's word embeddings with vectors trained on additional in-domain data.
To make this most effective, we make two other key changes to training.
First, we prevent embeddings from shifting during training.
Without this, the embedding space could become inconsistent as vectors for words seen in training shift while those for words seen only in the additional data stay the same.
Second, we do not tie the weights of the input embeddings and final output layer.
To understand the impact of these factors, we train models with every valid combination of weight tying, freezing, and pretraining.\footnote{Note, for frozen output embeddings the bias is not frozen.}

We experiment with \citet{merityRegOpt}'s AWD-LSTM -- a high-performing model that can be trained in under a day on a single GPU (without fine-tuning).
We train embeddings using GloVe on Gigaword.\footnote{
Embedding size 400 and rare word cutoff 5, the same as in the original AWD-LSTM model and GloVe respectively.
All other GloVe hyperparameters were set as specified in the original GloVe paper and trained using the released code.
}
For evaluation, we consider two versions of the Penn Treebank.
\emph{Std} is the standard version used in language modeling, with words of frequency less than five converted to UNK, all words lowercase, numbers replaced with a special symbol, and punctuation removed.
\emph{Rare} has the same pre-processing but without replacement of rare words.\footnote{
The script to generate our \emph{Rare} data from the LDC release is available at:  \url{http://jkk.name/emnlp20lm/}.
}

Table~\ref{tab:vary-train} shows the results, with icons to concisely describe the different configurations.\footnote{
Dice Icon by Andrew Doane from the Noun Project.
Fire and Snowflake Icons by Freepik from www.flaticon.com.
}
Looking first at the standard evaluation set, we can see the value of pretrained embeddings by considering pairs where the only difference is whether the embeddings are random or pretrained.
Pretrained embeddings are better in all but one case (comparing the fourth last and second last rows), and there the difference is only 0.5.
As for freezing the pretrained input embeddings, keeping all other aspects the same, it is always better to freeze them.

There are also four clear sections of performance in the table:
(a) frozen random output embeddings;
(b) frozen pretrained output embeddings;
(c) frozen random input embeddings;
(d) various configurations.
These results have an asymmetry.
Freezing the output embeddings consistently leads to poor performance, even with pretrained embeddings pretrained.
In contrast, freezing with pretrained input embeddings leads to some of the best results.
We expected freezing with random initialisation to perform poorly, but the drop is modest for input freezing and dramatic for output freezing.
This suggests that the two embedding matrices are serving different purposes in the model.
The results do support the practise of tying when the input embeddings are random, but the benefit is half as large when they are pretrained.

\begin{table}
    \hspace{-3mm}
    \scalebox{0.98}{
    \begin{tabular}{l c rl rl rr}
    \addlinespace[-\aboverulesep] 
     \cmidrule[\heavyrulewidth]{2-8}
        &         & \multicolumn{4}{c}{Embeddings} & \multicolumn{2}{c}{Dev PPL} \\
        & Tied    & \multicolumn{2}{c}{Input} & \multicolumn{2}{c}{Output} & Std & Rare \\
        \cmidrule(lr){1-1}
        \cmidrule[\lightrulewidth]{2-8}
        & \tied   & \frozen   & \dice  & \frozen   & \dice  & 680  & 1120 \\
        \multirow{3}{*}{(a)}
        & \untied & \frozen   & \dice  & \frozen   & \dice  & 680  & 1120 \\
        & \untied & \unfrozen & \dice  & \frozen   & \dice  & 680  & 431 \\
        & \untied & \unfrozen & \train & \frozen   & \dice  & 220  & 372 \\
        & \untied & \frozen   & \train & \frozen   & \dice  & 218  & 360 \\
        \cmidrule(lr){1-1}
        & \untied & \frozen   & \dice  & \frozen   & \train & 121  & 202 \\
        \multirow{3}{*}{(b)}
        & \untied & \unfrozen & \dice  & \frozen   & \train & 95.0 & 170 \\
        & \untied & \unfrozen & \train & \frozen   & \train & 91.3 & 147 \\
        & \tied   & \frozen   & \train & \frozen   & \train & 90.7 & 136 \\
        & \untied & \frozen   & \train & \frozen   & \train & 90.7 & 136 \\
        \cmidrule(lr){1-1}
        \multirow{2}{*}{(c)}
        & \untied & \frozen   & \dice  & \unfrozen & \dice  & 82.2 & 143 \\
        & \untied & \frozen   & \dice  & \unfrozen & \train & 81.4 & 142 \\
        \cmidrule(lr){1-1}
        & \untied & \unfrozen & \dice  & \unfrozen & \dice  & 65.3 & 120 \\
        & \untied & \unfrozen & \dice  & \unfrozen & \train & 64.1 & 113 \\
        & \untied & \unfrozen & \train & \unfrozen & \dice  & 62.5 & 105 \\
        \multirow{2}{*}{(d)}
        & \untied & \unfrozen & \train & \unfrozen & \train & 61.7 & 98.5 \\
        & \untied & \frozen   & \train & \unfrozen & \train & 61.6 & \textbf{97.1} \\
        & \tied   & \unfrozen & \dice  & \unfrozen & \dice  & 61.3 & 112  \\
        & \untied & \frozen   & \train & \unfrozen & \dice  & 61.1 & 98.1 \\
        & \tied   & \unfrozen & \train & \unfrozen & \train & \textbf{59.8} & 98.7 \\
        \cmidrule(lr){1-1}
     \cmidrule[\lightrulewidth]{2-8}
    \end{tabular}
    } \\
    
    \hspace{-3mm}
    \scalebox{0.95}{
    \setlength{\tabcolsep}{2pt}
    \begin{tabular}{ll c ll}
        \tied & = Tied parameters & & \untied & = Untied parameters \\
        \frozen & = Frozen in training & & \unfrozen & = Unfrozen in training \\
        \dice     & = Random init. & & \train  & = Pretrained init. \\
    \end{tabular}
    }
    \caption{\label{tab:vary-train}
    Perplexity on the PTB for all valid combinations of weight tying, freezing, and pretraining.
    Results are sorted by perplexity on Std and shown to three significant figures.
    }
\end{table}

For the dataset with rare words we see mostly the same trends.
The exception is the bottom six rows.
Once rare words are present, random initialisation of the input embeddings is considerably worse than pretraining (third last row).
Again, there is an asymmetry between input and output, with the top five models all using pretrained input embeddings, but only three of them using pretrained output embeddings.
Tying is also no longer the best approach, with the top three models not tying.
Our proposed approach, using pretrained untied embeddings and freezing the input, has the best results.

The only difference between Std and Rare is the lack of UNKs in Rare.
This impacts 5.1\% of tokens in the validation set (33\% of types).
While our pretrained embeddings do not cover all of these rare words, they do cover most.
The vocabulary from Gigaword that we build vectors for covers 99.5\% of the validation word tokens in Std (98\% of word types), and 98.8\% of the validation word tokens in Rare (84\% of word types).

\section{When \& Why Does Pretraining Help?}

To understand the strengths and limitations of this new approach, we consider a series of experiments, each probing a specific variable.
To simulate our target scenario, we use 44 million words of Wall Street Journal data from the North American News Corpus \cite[NANC,][]{nanc}.
This provides enough data for pretraining, training, validation, and test sets all in the exact same domain (not even varying the newspaper).
We apply similar pre-processing as in the previous section, but break the data down into articles rather than sentences and keep rare words.

We compare the six best configurations from Table~\ref{tab:vary-train}.
In all cases, output embeddings are not frozen, so we leave out the symbol.
We use only one symbol for pretraining/random because both embeddings are the same in most cases.
The exceptions have \halftrain\xspace to indicate pretrained input and random output.

\noindent
{\setlength{\tabcolsep}{3pt}
\begin{tabular}{lp{5.8cm}}
    \tied\xspace \unfrozen\xspace \dice & Standard approach. \\
    \untied\xspace \unfrozen\xspace \halftrain & Our approach, but with random output embeddings and without freezing.\\
    \tied\xspace \unfrozen\xspace \train & Standard approach + pretraining. \\
    \untied\xspace \unfrozen\xspace \train & Our approach, but without freezing. \\
    \untied\xspace \frozen\xspace \train & Our approach. \\
    \untied\xspace \frozen\xspace \halftrain & Our approach, but with random output embeddings. \\
\end{tabular}
}

\tightparagraph{Other Domains Show the Same Pattern.}
First we consider varying the domain to make sure this is not an artifact of news data.
Table~\ref{tab:vary-domain} shows results on Covid-19 research \cite{cord}, Ubuntu IRC chat \cite{acl19disentangle}, Reddit, and Wikipedia, tokenised with either Scispacy \cite{scispacy} or Stanza \cite{stanza}.
Pretraining consistently helps, while freezing is best on all but Wikipedia.
Our approach is consistently either the best or very close to the best.

\tightparagraph{The Improvement is Due to Rare Words.}
To probe the impact of rare words, we explore replacing them with UNK (using the same UNK symbol as used in embedding pretraining).
We consider four variations, each constructed in two steps.
First, we make a list of the words in the original training set and how many times each one occurs.
Second, we make modified versions of the training and validation sets, replacing words with UNK if their count in our list is lower than K.
For this step, any word that does not appear in our list is treated as if it has a count of zero.
We consider K = 0, 1, 2 and 5.
K is 0 for all other experiments in this section, which means that no words are replaced with UNK.
When K is 1, 2, and 5, the introduction of UNKs means all words in the validation set are seen during language model training.

\begin{table}
    \centering
    \scalebox{\tablescale}{
    \begin{tabular}{l ccccc}
        \toprule
        Train & \multicolumn{5}{c}{Domain} \\
        Config & NANC & Cord & IRC & Reddit & Wiki \\
        \midrule                      
        \tied\xspace   \unfrozen\xspace \dice  &  106 & 135 & 41.3 & 186 & 206 \\
        \untied\xspace \unfrozen\xspace \halftrain & 103 & 125 & 41.1 & 166 & 174 \\
        \tied\xspace   \unfrozen\xspace \train & 97.2 & 121 & 39.8 & 154 & 142 \\
        \untied\xspace \unfrozen\xspace \train & 95.7 & 111 & 39.2 & 152 & \textbf{141} \\
        \untied\xspace \frozen\xspace   \train & 90.8 & \textbf{109} & \textbf{37.3} & \textbf{146} & 144 \\
        \untied\xspace \frozen\xspace   \halftrain & \textbf{90.5} & 112 & 37.6 & 152 & 161 \\
        \bottomrule
    \end{tabular}
    }
    \caption{\label{tab:vary-domain}
    Results for various domains.
    All other results in this section are for NANC.
    }
\end{table}

\begin{table}
    \centering
    \scalebox{\tablescale}{
    \begin{tabular}{l rrrr}
        \toprule
        Train & \multicolumn{4}{c}{Frequency Cutoff} \\
        Config & 0 & 1 & 2 & 5 \\
        \midrule                      
        \tied\xspace   \unfrozen\xspace \dice  &  106 &  106 & 70.6 & 55.4 \\
        \untied\xspace \unfrozen\xspace \halftrain & 103 & 104 & 72.5 & 56.8 \\
        \tied\xspace   \unfrozen\xspace \train & 97.2 & 99.9 & 68.1 & 54.1 \\
        \untied\xspace \unfrozen\xspace \train & 95.7 & 97.8 & 70.2 & 56.0 \\
        \untied\xspace \frozen\xspace   \train & 90.8 & 92.1 & 66.5 & 54.5 \\
        \untied\xspace \frozen\xspace   \halftrain & \textbf{90.5} & \textbf{91.5} & \textbf{65.8} & \textbf{54.0} \\
        \midrule
        UNK Types Dev & 0\% & 13\% & 21\% & 33\% \\
        UNK Tokens Dev & 0\% & 2.3\% & 3.4\% & 5.5\% \\
        UNK Types Train & 0\% & 0\% & 40\% & 68\% \\
        UNK Tokens Train & 0\% & 0\% & 1.4\% & 4.1\% \\
        \bottomrule
    \end{tabular}
    }
    \caption{\label{tab:vary-data}
    Varying the minimum frequency to not be converted into an UNK.
    The top half shows language model perplexity.
    The bottom half shows the percentage of word tokens and types that are replaced with UNK in each case.
    }
\end{table}

\begin{table}
    \centering
    \scalebox{\tablescale}{
    \begin{tabular}{lrrrr|rr}
        \toprule
        & \multicolumn{2}{c}{Train}
        & \multicolumn{2}{c}{Pretrain}
        & \multicolumn{2}{c}{Train in Pre} \\
        Dataset & Type & Tok & Type & Tok & Type & Tok \\
        \midrule
        PTB    &  73 & 5.3 &  77 & 0.11 &  14 & 1.3  \\
        NANC   &  71 & 4.8 &  63 & 0.49 &  13 & 0.63 \\
        Sci    &  78 & 6.3 &  85 & 1.2  &  23 & 1.6  \\
        IRC    &  83 & 4.2 &  90 & 1.3  &  37 & 1.4  \\
        Reddit &  81 & 6.1 &  86 & 0.69 &  15 & 0.71 \\
        Wiki   &  78 & 7.3 &  78 & 0.36 & 5.6 & 0.43 \\
        \bottomrule
    \end{tabular}
    }
    \caption{\label{tab:word-freq}
    Percentage of word types and tokens that occur five times or fewer in each dataset.
    The last two columns are the percentage of types/tokens in the training set that occur five or fewer times in the pretraining set.
    For PTB the pretraining set is Gigaword (as used in Table~\ref{tab:vary-train}).
    }
\end{table}

Table~\ref{tab:vary-data} shows a clear trend: the benefit of our approach grows as more rare words are present (i.e., K is smaller).
Note, it may seem odd that perplexity is higher when K=1 than when K=0 since we have removed rare words.
This is probably because when K is 1 there are UNKs in the validation set but not in the language model training set.

Table~\ref{tab:word-freq} shows statistics about rare words in the datasets.
71-83\% of word types in the training sets occur fewer than five times, but most of these appear frequently in the pretraining sets (compare the first column with the second last column).
The same pattern occurs for word tokens.
Comparing the statistics for the training set and the pretraining set, the percentage of rare word types is fairly consistent while the percentage of rare tokens consistently goes down.

\tightparagraph{Pretraining Data Needs to be from a Similar Domain.}
We would expect that the effectiveness of pretraining will depend on how similar the data is.
Table~\ref{tab:vary-pretrain} shows results with different embeddings, and indicates the number of words used in pretraining.
We see that the value of additional data depends on the domain.
Gigaword is also news text and is able to improve performance.
The larger GloVe datasets use Wikipedia and CommonCrawl data, which is a poorer match and so does not improve performance.
For GloVe we did have to change the embedding dimensions from 400 to 300, which may impact performance slightly.

\begin{table}
    \centering
    \scalebox{\tablescale}{
    \begin{tabular}{l ccccc}
        \toprule
         & \multicolumn{4}{c}{Pretraining Source} \\
        Train & NANC & Gigaword & \multicolumn{2}{c}{GloVe} \\
        Config & 43M & 5B & 6B & 42B \\
        \midrule                      
        \tied\xspace   \unfrozen\xspace \train & 97.2 & 90.9 & 93.1 & 93.3 \\
        \untied\xspace \unfrozen\xspace \halftrain  & 103 & 99.3 & 98.3 & 99.5 \\
        \untied\xspace \unfrozen\xspace \train   & 95.7 & \textbf{90.0} & 91.2 & 93.6 \\
        \untied\xspace \frozen\xspace \train     & 90.8 & 90.6 & 91.1 & 91.5 \\
        \untied\xspace \frozen\xspace   \halftrain  & 90.5 & 90.7 & 90.7 & 91.9 \\
        \bottomrule
    \end{tabular}
    }
    \caption{\label{tab:vary-pretrain}
    Varying similarity and size of pretraining data.
    Dataset size is shown below the name of each dataset.
    }
\end{table}

\tightparagraph{The Effect Persists When Language Model Training Data is Increased.}
So far we have only used the additional in-domain data for pretraining.
In this experiment, we expand the training set for the language model.
We try two variations, one where the data is an exact domain match (NANC) and one where it is also news, but from different newspapers and from a different year (Gigaword).
Table~\ref{tab:lm-data} shows that as we increase the amount of data our approach and the variant with random output embeddings continue to do best, but the margin shrinks between them and the standard approach.
Note, however, that these results are with hyperparameters tuned for the baseline configuration.
With tuning the 0.7 gap between our proposal and the baseline for 4xNANC widens to 6.6.

\begin{table}
    \centering
    \scalebox{\tablescale}{
    \begin{tabular}{l rrr | rrr}
        \toprule
        Train & \multicolumn{3}{c}{NANC WSJ} & \multicolumn{3}{c}{Gigaword} \\
        Config &   1x &   2x &   4x &   1x &   2x &   4x \\
        \midrule                      
        \tied\xspace   \unfrozen\xspace \dice  &  106 & 81.0 & 67.5 & 106 & 92.5 & 86.3 \\
        \untied\xspace \unfrozen\xspace \halftrain & 103 & 83.3 & 68.7 & 99.3 & 91.7 & 87.2 \\
        \tied\xspace   \unfrozen\xspace \train & 97.2 & 80.4 & 67.8 & 90.9 & 88.6 & 85.7 \\
        \untied\xspace \unfrozen\xspace \train & 95.7 & 80.0 & 68.1 & \textbf{90.0} & 86.4 & 85.5 \\
        \untied\xspace \frozen\xspace   \train & 90.8 & 73.7 & 66.8 & 90.6 & 84.8 & \textbf{82.5} \\
        \untied\xspace \frozen\xspace   \halftrain & \textbf{90.5} & \textbf{72.9} & \textbf{66.1} & 90.7 & \textbf{83.8} & 83.7 \\
        \bottomrule
    \end{tabular}
    }
    \caption{\label{tab:lm-data}
    Expanding the language model training set.
    }
\end{table}

\begin{figure}
    \centering
    \includegraphics[width=\linewidth]{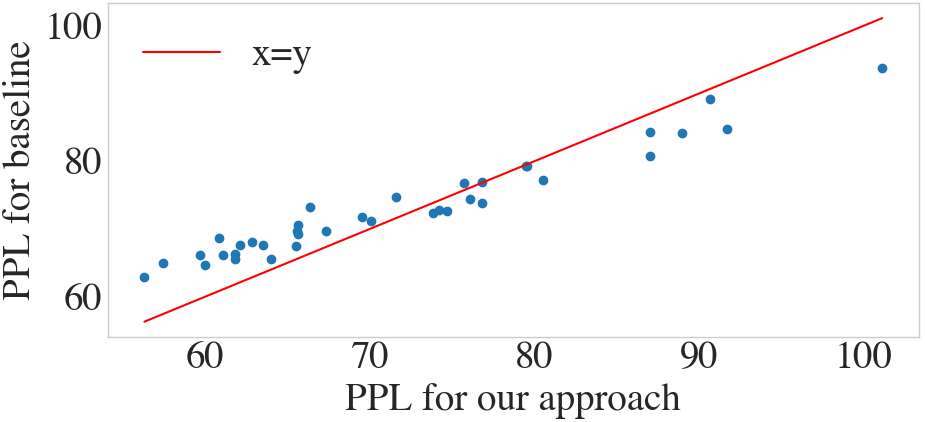}
    \caption{\label{fig:tuning}
    Hyperparameter search results with one point for each configuration.
    The line separates where our approach is better (left) or worse (right).
    }
\end{figure}

\tightparagraph{Hyperparameter Tuning Further Improves Results.}
All of the previous experiments were slightly tipped in favour of the baseline as we used the hyperparameters from \citet{merityRegOpt}.
We do not have the resources to tune for every condition, so instead we focus on a final set of experiments with the 4xNANC condition from Table~\ref{tab:lm-data}.
We run 37 configurations with randomly sampled hyperparameters, using the same configurations for the baseline and our proposed approach (see the supplementary material for details).
Figure~\ref{fig:tuning} shows that our approach is even stronger after tuning, with a score that is 6.6 better than the baseline.
Comparing the baseline and tuned hyperparameters, some shifted substantially more than others:
the learning rate was halved;
word dropout was halved;
and the number of layers was increased from 3 to 4.
The other parameters shifted by 15-30\%.

\tightparagraph{Test Results Confirm Our Observations.}
Using the best configuration we train the baseline and our proposed approach using 8xNANC (the most our GPU could support).
We compare to an n-gram language model trained on all of the NANC data \cite{heafield-etal-2013-scalable}, and a transformer based model trained on a massive dataset, GPT-2 \cite{gpt2}.
While GPT-2 cannot be retrained in a low-compute scenario, it can be used.
We compare to GPT-2 without fine-tuning.
We evaluate byte-pair encoding (BPE) separately because with BPE tokenisation models have additional information when predicting the second or later piece of a token \cite{sha-rnn}.

\begin{table}
    \centering
    \scalebox{\tablescale}{
    \begin{tabular}{l rr|rr}
        \toprule
                & \multicolumn{2}{c}{Words} & \multicolumn{2}{c}{BPE} \\
        Model   & Dev & Test & Dev & Test \\
        \midrule
        N-Gram                                              & 92.3 & 95.0 & 56.7 & 55.3 \\
        GPT-2 (112m)                                        &    - &    - & 46.4 & 43.8 \\
        Baseline AWD-LSTM                                   & 52.8 & 53.5 & 37.8 & 36.7 \\
        Our approach                                        & 49.0 & 49.4 & 38.3 & 37.2 \\
        GPT-2 (774m)                                        &    - &    - & 32.5 & 33.7 \\
        \bottomrule
    \end{tabular}
    }
    \caption{\label{tab:final}
    Final results, training with 8xNANC.
    }
\end{table}

Table~\ref{tab:final} shows that for word-level prediction, our approach improves over the baseline and an n-gram language model.
BPE breaks up rare words, leading to no improvement over the baseline and while we do better than the 112m parameter GPT-2, we do not do as well as the 774m parameter one (both untuned).
Overall, this indicates that for users who require word-level scores and have limited computational resources our approach is an effective way to use additional data when training LSTM language models.

\section{Related Work} \label{sec:related-work}

\tightparagraph{Embedding Tying.}
Tying input and output matrices has consistently increased performance while reducing the number of model parameters \cite{press-wolf-2017-using,Inan2017TyingWV}.
The improvement is thought to be because otherwise only one input embedding is updated each step and the gradient has to propagate a long way through the model to reach it.
Subsequent work has explored more advanced forms of tying, recognising that the role of the input and output matrices are not exactly the same \cite{pappas-etal-2018-beyond}.
This asymmetry has been found in the actual embedding spaces learned and shown to have a negative effect on performance \cite{Gao:ICLR:19,Demeter2020StolenPA}.
These observations match the patterns we observe and provide theoretical justification for not tying when possible.

\tightparagraph{In-Domain Data Pretraining and Freezing.}
Word vectors are frequently used in downstream tasks and recent work has shown that their effectiveness depends on domain similarity \cite{peters-etal-2019-tune,Arora2020ContextualEW}
For language modeling, \citet{kocmi-bojar-2017-exploration} explored random and pretrained embeddings and found improvements, but did not consider tying and freezing.
In-domain data is also useful for continuing to train contextual embedding models before fine-tuning \cite{Gu2020TrainNE,Gururangan2020DontSP}, and for monolingual pretraining in machine translation \cite{neishi-etal-2017-bag,qi-etal-2018-pre,Artetxe2018UnsupervisedNM}.
This matches our observations, but does not cover the interactions between freezing and tying we consider.

\tightparagraph{Handling Rare Words.}
These remain challenging even for large transformer models \cite{Schick2019BERTRAMIW}.
Recent work has explored copying mechanisms and character based generation \cite{kawakami-etal-2017-learning}, with some success.
These ideas are complementary to the results of our work, extending coverage to the open vocabulary case.
Due to space and computational constraints we only consider English.
For other languages, inflectional morphology and other factors may impact the effectiveness of our approach \cite{shareghi-etal-2019-show,cotterell-etal-2018-languages}.
Our work is also complementary to concurrent work on producing rare words as output \cite{pappas-2020}.

\tightparagraph{Language Model Types.}
We focus on a single model type for computational budget reasons.
We chose an LSTM because while transformer based models such as GPT-2 now dominate transfer learning, LSTMs continue to be competitive in language modeling \cite{Du2020ExploitingSS,Li2020LearningAF,Melis2018OnTS,merityRegOpt}.
Our ideas are orthogonal to this prior work and our findings may apply to transformers as well, but confirming that would require additional experiments.

\section{Conclusion}

Initialising embeddings with vectors trained on in-domain data can improve performance by providing better representations for rare words.
This effect persists even as more in-domain data is used to train the language model.
Our work also suggests that standard model components like embedding tying should be retested as we continue to explore the space of language modeling.

\section*{Acknowledgements}

We would like to thank Greg Durrett for helpful feedback on an earlier draft of this paper and the anonymous reviewers for their helpful suggestions.
This material is based in part on work supported by DARPA (grant \#D19AP00079), Bloomberg (Data Science Research Grant), the National Science Foundation (grant \#1815291), and the John Templeton Foundation (grant \#61156).

\bibliography{emnlp2020}
\bibliographystyle{acl_natbib}

\end{document}